\PassOptionsToPackage{unicode}{hyperref}
\PassOptionsToPackage{hyphens}{url}
\documentclass[
]{article}
\usepackage{lmodern}
\usepackage{amssymb,amsmath}
\usepackage{ifxetex,ifluatex}
\ifnum 0\ifxetex 1\fi\ifluatex 1\fi=0 
  \usepackage[T1]{fontenc}
  \usepackage[utf8]{inputenc}
  \usepackage{textcomp} 
\else 
  \usepackage{unicode-math}
  \defaultfontfeatures{Scale=MatchLowercase}
  \defaultfontfeatures[\rmfamily]{Ligatures=TeX,Scale=1}
\fi
\IfFileExists{upquote.sty}{\usepackage{upquote}}{}
\IfFileExists{microtype.sty}{
  \usepackage[]{microtype}
  \UseMicrotypeSet[protrusion]{basicmath} 
}{}
\makeatletter
\@ifundefined{KOMAClassName}{
  \IfFileExists{parskip.sty}{%
    \usepackage{parskip}
  }{
    \setlength{\parindent}{0pt}
    \setlength{\parskip}{6pt plus 2pt minus 1pt}}
}{
  \KOMAoptions{parskip=half}}
\makeatother
\usepackage{xcolor}
\IfFileExists{xurl.sty}{\usepackage{xurl}}{} 
\IfFileExists{bookmark.sty}{\usepackage{bookmark}}{\usepackage{hyperref}}
\hypersetup{
  hidelinks,
  pdfcreator={LaTeX via pandoc}}
\usepackage{graphicx}
\makeatletter
\def\maxwidth{\ifdim\Gin@nat@width>\linewidth\linewidth\else\Gin@nat@width\fi}
\def\maxheight{\ifdim\Gin@nat@height>\textheight\textheight\else\Gin@nat@height\fi}
\makeatother
\setkeys{Gin}{width=\maxwidth,height=\maxheight,keepaspectratio}
\makeatletter
\def\fps@figure{htbp}
\makeatother
\usepackage[final]{neurips_2021}
\usepackage[utf8]{inputenc}
\usepackage[T1]{fontenc}
\usepackage{hyperref}
\usepackage{url}
\usepackage{booktabs}
\usepackage{amsfonts}
\usepackage{nicefrac}
\usepackage{microtype}
\usepackage{wrapfig, blindtext}
\usepackage{xcolor}

\usepackage{tcmmacros}
\newlength{\cslhangindent}
  {\setlength{\parindent}{0pt}%
  \everypar{\setlength{\hangindent}{\cslhangindent}}\ignorespaces}%
  {\par}

\begin{document}

\title{DeepSITH: Efficient Learning via Decomposition of What and When Across Time Scales}

\author{%
Brandon G. Jacques \\
Department of Psychology \\
University Of Virginia \\
\texttt{bgj5hk@virginia.edu} \\
\And
Zoran Tiganj \\
Department of Computer Science \\
Indiana University \\
\texttt{ztiganj@iu.edu} \\
\AND
Marc W. Howard \\
Department of Psychological and Brain Sciences \\
Boston University \\
\texttt{marc777@bu.edu} \\
\And
Per B. Sederberg \\
Department of Psychology \\
University of Virginia \\
\texttt{pbs5u@virginia.edu} \\
}

\maketitle

\begin{abstract}
Extracting temporal relationships over a range of scales is a hallmark of
human perception and cognition---and thus it is a critical feature of machine
learning applied to real-world problems.  Neural networks are either plagued
by the exploding/vanishing gradient problem in recurrent neural networks
(RNNs) or must adjust their parameters to learn the relevant time scales
(e.g., in LSTMs). This paper introduces DeepSITH, a deep network comprising
biologically-inspired Scale-Invariant Temporal History (SITH) modules in
series with dense connections between layers. Each SITH module is simply a
set of time cells coding what happened when with a geometrically-spaced set of
time lags.  The dense connections between layers change the definition of what
from one layer to the next.  The geometric series of time lags implies that
the network codes time on a logarithmic scale, enabling DeepSITH network to
learn problems requiring memory over a wide range of time scales. We compare
DeepSITH to LSTMs and other recent RNNs on several time series prediction and
decoding tasks. DeepSITH achieves results comparable to state-of-the-art
performance on these problems and continues to perform well even as the delays
are increased.
\end{abstract}

\hypertarget{introduction}{%
\section{Introduction}\label{introduction}}

The natural world contains structure at many different time scales.
Natural learners can spontaneously extract meaningful information from a
range of nested time scales allowing a listener of music to appreciate
the structure of a concerto over time scales ranging from milliseconds
to thousands of seconds. Recurrent neural networks (RNNs) enable
information to persist over time and have been proposed as models of
natural memory in brain circuits \cite{Rajan.etal.2016}. For decades,
long-range temporal dependencies have been recognized as a serious
challenge for RNNs \cite{Mozer.1991}. This problem with RNNs is fundamental,
arising from the exploding/vanishing gradient problem \cite{Bengio.etal.1994, Pascanu.etal.2012}. The
importance of long-range dependencies coupled with the difficulties with RNNs
has led to a resurgence of interest in long short-term memory networks (LSTMs)
over the last several years.

LSTMs, however, are neurobiologically implausible and, as
a practical matter, tend to fail when time scales are very large
\cite{Hochreiter.Schmidhuber.1997, Li.etal.2019}. More recent
approaches attempt to solve the problem of learning across multiple
scales by constructing a scale-invariant memory. For example, Legendre
Memory Units (LMU) are RNNs that utilize a specialized weight
initialization technique that theoretically guarantees the construction
of long time-scale associations \cite{Voelker.etal.2019}. LMUs construct a
memory for the recent past using Legendre polynomials as basis
functions. In a different approach, the Coupled oscillatory Recurrent
Neural Network (coRNN) \cite{Rusch.Mishra.2020} treats each internal node
as a series of coupled oscillators which has the benefit of
orthogonalizing every discrete moment in time. In this paper, we
introduce a novel approach to machine learning problems that depends on
long-range dependencies inspired by recent advances in the neuroscience
of memory.

\begin{wrapfigure}{l}{0.55\linewidth}
\hypertarget{fig:model_config}{%
\centering
\includegraphics{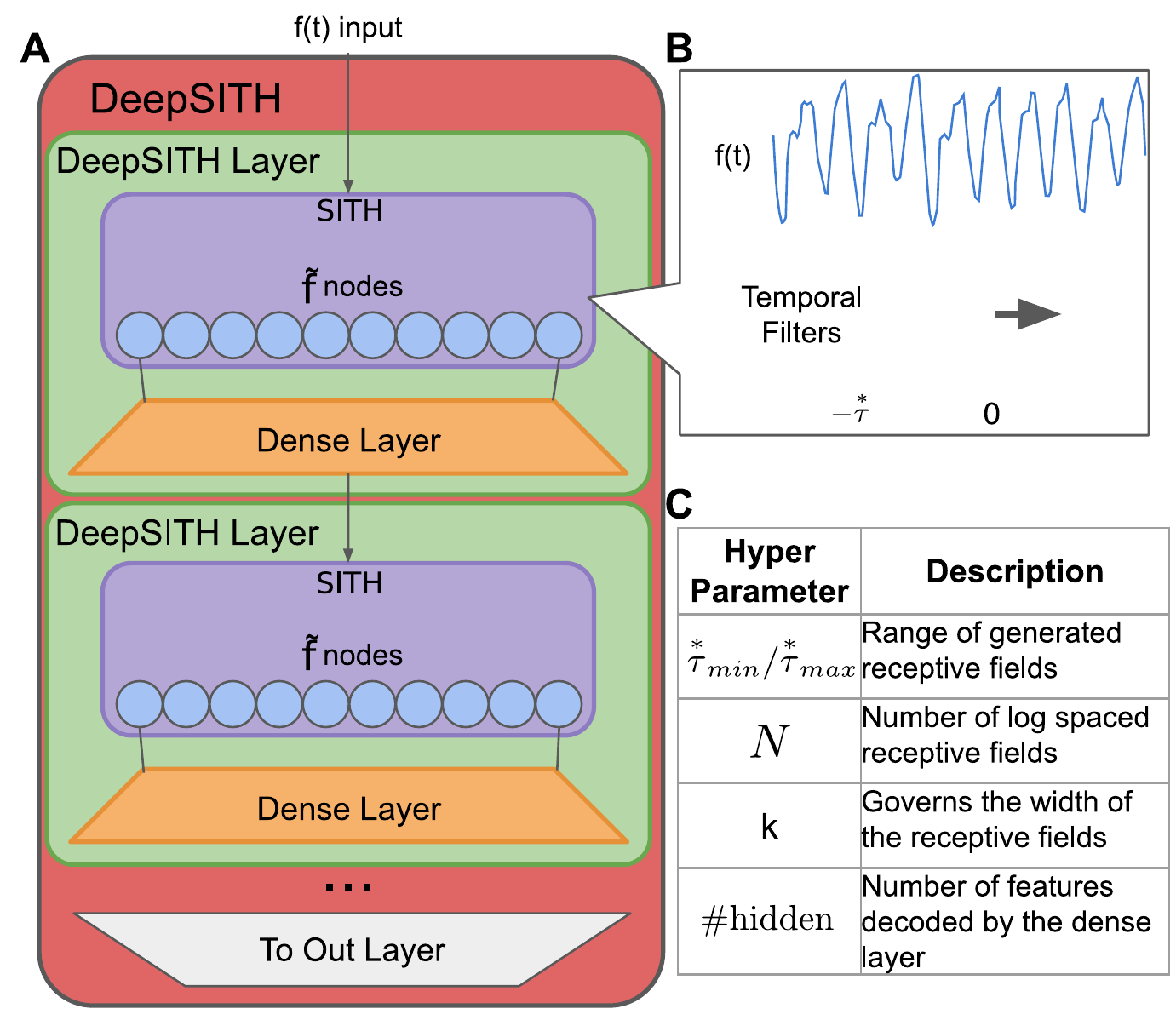}
\caption{\emph{Structure of the DeepSITH network.} \textbf{A}: A diagram
of the DeepSITH network, depicting an example with two layers.
\textbf{B}: The input signal \(f(t)\) is convolved with the
precalculated temporal filters to produce an output of the
\(\overset{*}{\tau}\) nodes. \textbf{C}: A table outlining the
parameters that need to be specified for each DeepSITH layer. Typically,
we recommend that \numtaustar{}, \(\overset{*}{\tau}_{min}\), and hidden
size remain constant across layers, but \(\overset{*}{\tau}_{max}\)
should increase logarithmically, and k is calculated via the formula
outlined in the text. The \(\#hidden\) parameter dictates the output of
the dense layers, constant across layers, combining both what and when
information into new features.}\label{fig:model_config}
}
\end{wrapfigure}

In neuroscience, it has become increasingly clear that the brain makes
use of a memory for the recent past that 1) contains information about
both the time and identity of past events 2) represents the time of past
events with decreasing accuracy for events further in the past and 3)
represents the past over many different time scales. So-called ``time
cells'' with these properties have been observed in many different brain
regions in multiple mammalian species in a variety of behavioral tasks
(see \cite{Eichenbaum.2017}, and \cite{Howard.2018} for reviews).
Recent evidence suggests that the brain constructs temporal basis functions
over logarithmic time \cite{GuoEtal21, CaoEtal21}.
These neuroscience
findings were anticipated by computational work that describes a method
for computing a scale-invariant temporal history \cite{Shankar.Howard.2012}
 and consistent with a long tradition in human memory research
noting the scale-invariance of behavioral memory \cite{Pashler.etal.2009}.
Here, we introduce a Deep Scale-Invariant Temporal History (DeepSITH)
network. This network consists of a series of layers. Each layer
includes a biologically-inspired scale-invariant temporal history
(SITH)---effectively a set of time cells
that remembers what happened when---the history of its inputs---along a
logarithmically-compressed time axis. A dense layer with learnable weights
connects each layer to the
next (see Figure \ref{fig:model_config}), transforming temporal relationships
into new features for the next layer.
The next layer also codes for what happened when but the meaning of ``what''
changes from one layer to the next.  We compare DeepSITH to an
LSTM, LMU and coRNN on a set of time series prediction and decoding
tasks designed to tax the networks' ability to learn and exploit
long-range dependencies in supervised learning situations.

\hypertarget{scale-invariant-temporal-history}{%
\subsection{Scale-invariant temporal
history}\label{scale-invariant-temporal-history}}

Each DeepSITH layer includes a scale-invariant temporal history (SITH)
layer. Given a time series input \(f(t)\), the state of the memory at
time \(t\) is denoted \(\ftilde(t,\taustar)\). This memory approximates
the past in that \(\ftilde(t,\taustar)\) is an approximation of
\(f(t-\taustar)\). At each moment \(t\), \(\ftilde(t,\taustar)\) is
given by
\begin{equation}\ftilde (t,\taustar) = \taustar^{-1} \int_{t’=t}^{-\infty} \Phi_k\left(\frac{t’}{ \taustar}\right) \  f \left(t - t’\right) dt’,\label{eq:ftilde}\end{equation}
where the scale-invariant filter \(\Phi(x)\) is just a gamma function
\begin{equation}{\Phi_k}(x) \propto ( x )^k\ e^{-kx}.\label{eq:phidef}\end{equation}

\begin{figure}[t!]
\hypertarget{fig:std_explanation}{%
\centering
\includegraphics[width=0.9\textwidth]{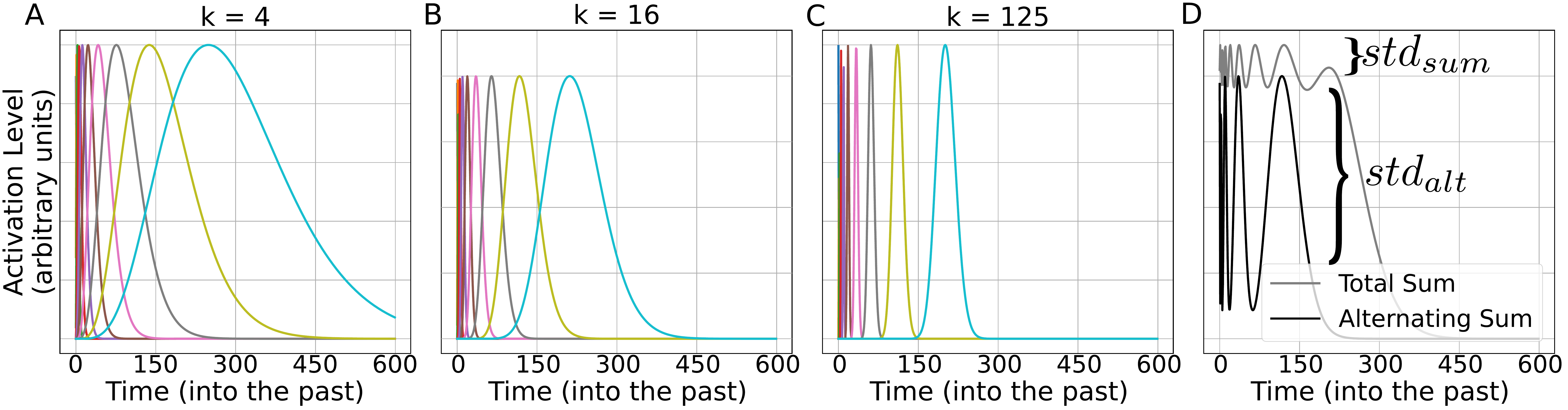}
\caption{\emph{SITH temporal filters and selection of \(k\).} Curves in
panels \textbf{A}, \textbf{B}, and \textbf{C} show impulse responses of
10 SITH filters, with each panel corresponding to a different value of
parameter \(k\), which controls the coefficient of variation of the
filters (larger \(k\) results in narrower filters). Note that the
filters are scale-invariant: the coefficient of variation is
proportional to the peak time. The filters in this figure have been
normalized for easier visualization and for our calculation of the
optimal \(k\). In the experiments the amplitude will decay as a
power-law function of time, such that the area under each filter is the
same. In panel \textbf{B} we applied an automated method to optimize the
selection of \(k\) (here, \(k=16\)), such that the filters were
overlapping, but not so much that there was redundant information.
\textbf{D} illustrates the optimization approach minimizing the ratio
between the standard deviation across time of the sum of all of the
filters and the standard deviation of the sums of every other filter
(Please see the supplemental information for more
detail.)}\label{fig:std_explanation}
}
\end{figure}

The function \(\Phi_k(x)\) defined in Eq. \ref{eq:phidef} describes a
unimodal impulse response function that peaks at \(\taustar\). The width of
the impulse response function is controlled by \(k\); higher values of
\(k\) result in sharper peaks (Fig. \ref{fig:std_explanation}). Because
the filter in Eq. \ref{eq:ftilde} is a function of \(t/\taustar\),
\(\ftilde(t, \taustar)\) samples different parts of \(f(t’ < t)\) with
the same \emph{relative} resolution \cite{Shankar.Howard.2012}.

In implementing this formal approach, the variable \(\taustar\) must be
mapped onto a population of units. Noting the scale-covariance of
\(\ftilde (t, \taustar)\) one obtains constant resolution per unit if
the \(\taustar\) of the \(i\)th unit is chosen as
\[ \taustar_i = \taustar_{\textnormal{min}}\left(1+c\right)^i \] This means
that \(\taustar\) values are sampled more densely for values close to
the present \(\taustar=0\) and less densely for time points further in
the past. This property is shared with populations of time cells in the
brain.  More precisely, the expression for $\taustar_i$ above implies that the
$\taustar$ values are evenly spaced on a logarithmic axis.  Recent evidence
suggests that the brain codes time on a logarithmic axis as well
\cite{GuoEtal21,CaoEtal21}.

Equation \ref{eq:ftilde} appears to require access to the entire temporal
history to construct the representation \(\ftilde(t, \taustar)\); in the
experiments described below, we used numerical convolution of
\(f(t’ < t)\) and \(\Phi_k\) to construct \(\ftilde(t,\taustar)\). However, it
is possible to construct \(\ftilde(t, \taustar)\)
without remembering the entire history.  For instance, one might first
construct the real Laplace transform of \(f(t’ < t)\), \(F(s)\), using a
time-local differential equation as
\begin{equation}\frac{d{F}(t, {s})}{dt} = -s {F}(t, s) + f(t).\label{eq:lap_update}\end{equation}
The filter \(\Phi_k\) described above is the analytic result for the
Post approximation for the inverse Laplace transform, which can be
approximated with a linear operator \(\Lk\):
\begin{equation}\ftilde(t, \taustar)= \Lk F(t, s) \equiv C_k {s}^{k+1} \frac{d^k}{ds^k} {F}\left(t,s\right)\label{eq:inv_lap}\end{equation}
with the mapping \(s = k/\taustar\). Defining \(\ftilde\) in this way
gives Eq. \ref{eq:ftilde} as a solution.  There are other ways one could
construct $\ftilde$ without remembering the entire history
\cite{deVrPrin92,Lind16}.

Note that Eq. \ref{eq:inv_lap} does not require storing the history of
\(f\) to construct the real Laplace transform of that history.  It is
possible to sample \(s\) logarithmically as well as \(\taustar\),
resulting in exponential memory savings. However, in the experiments
presented here we used numerical convolution to avoid errors that arise
from approximating derivatives with large values of \(k\) \cite{Gosmann.2018}.
To distinguish this from prior machine learning work that used a
direct implementation of \(\Lk\), we refer to this implementation of
scale-invariant temporal history as iSITH.

It is perhaps worth noting that the Laplace transform in Eq.
\ref{eq:lap_update} can be understood as an RNN with a diagonal
connectivity matrix that does not change with learning. Similarly,
\(\ftilde\) can also be understood as an RNN with fixed weights, that is, a
reservoir computer.  The reservoir has a very specific form, however.  The
reservoir is chosen such that the eigenvalues are in geometric series and the
eigenvectors are translated versions of one another \cite{LiuHowa20}.   Taken
together these two properties correspond to the statement that $\ftilde$
represents what happened when as a function of log time.  DeepSITH can thus be
understood simply as a deep reservoir computer with this specialized form for
the reservoir.

\hypertarget{deepsith-a-deep-neural-network-using-neurally-plausible-representations-of-time}{%
\subsection{DeepSITH: A Deep Neural Network Using Neurally Plausible
Representations of
Time}\label{deepsith-a-deep-neural-network-using-neurally-plausible-representations-of-time}}

A DeepSITH network consists of a series of DeepSITH layers. At the input
stage of the \(i\)th DeepSITH layer, a SITH representation is constructed
for each of the \numfeat{i} input features.
The number of units in this SITH representation is
equal to the number of input features \numfeat{i} times the number
of \(\taustar\)s, \numtaustar{i}. At the output stage of each DeepSITH
layer, the SITH representation is fed through a dense layer with
modifiable weights \(\mathbf{W}^{(i)}\) and a ReLU activation function
\(g(.)\):
\begin{equation}\mathbf{o}(t) = g\left[\mathbf{W}^{(i)} {\ftilde}(t, \taustar)\right].\end{equation}
The weight matrix \(\mathbf{W}^{(i)}\) connects the output of layer
\(i\) to the input of layer \(i+1\). \(\mathbf{W}^{(i)}\) has
\(\numfeat{i+1} \times (\numfeat{i} \times \numtaustar{i})\) entries. In
the experiments examined here, the input to the first layer is one- or
two-dimensional, but the number of features is greater for subsequent
layers.

One final dense linear layer converts the output from the final DeepSITH
layer to the dimensionality required for the specific problem. A diagram
depicting this network is shown in Figure \ref{fig:model_config}.A. We provide
all the code for DeepSITH and the subesquent analysis in our github \href{https://github.com/compmem/DeepSITH}{here}.

\hypertarget{parameterization-of-deepsith}{%
\subsection{Parameterization of
DeepSITH}\label{parameterization-of-deepsith}}

There are a few hyper-parameters that need to be specified to optimize
the performance of a DeepSITH network. A
network with three layers proved successful for most of the
experiments here, but four layers were used for Mackey-Glass and Hateful-8
described below. Layer-specific parameters are \(\taustar_{\textnormal{max}}\)
and number of \(\taustar\)s, \(\numtaustar{i}\). We set \(\taustar_{max}\) to
increase geometrically from layer to layer, and \(\taustar_{min}\) was set to 1 \(\Delta t\) for all
problems.  The number of \(\taustar\)s, \(\numtaustar{i}\) was constant across
layers; we found that values from 10-30 were roughly equivalent.

The value of \(k\) was chosen to be dependent on the values of \(\taustar_{max}\)
and \(\numtaustar{i}\) (see Figure \ref{fig:std_explanation}.D). The density of the
centers of the temporal filters are controlled by the value $c$, where
\(c = \left(\taustar_{max}/\taustar_{min}\right)^{1/\numtaustar{}} + 1\). The rate in which
width of the filters increase as a function of \(\taustar\) is dependent on \(k\).
We choose \(k\) in order to minimize the ratio of the standard deviation of the
sum of all the filters, \(std_{all}\), to the standard deviation of the sum of
alternating filters, \(std_{alt}\). This minimization is so that, for a given \(c\),
we do not over represent the past in our iSITH representation, nor do we
construct temporal filters in iSITH that have too much overlap. More detail on
this minimization is provided in the supplemental sections.

The final key hyper-parameters when setting up a DeepSITH network are
the sizes of the dense layers. This decision depends largely on the
number of input features and the expected complexity of the input. The
number of output features from each dense layer, or the hidden size, is
an estimate of how many unique temporal associations can be extracted at
each moment from the history of input features decomposed via the SITH
layer. In order to keep the network size small, this hidden size should
be kept relatively small (less than 100 nodes), and we have found in practice
that it can be consistent across layers after the first layer. In the cases where an
input signal is sufficiently complex in the breadth of temporal dynamics
that need to be encoded, a larger hidden size may be required.

Table \ref{tab:model_params} summarizes the hyperparameters used in the experiments.

\hypertarget{experiments}{%
\section{Experiments}\label{experiments}}

We compared the DeepSITH network to previous RNNs---LSTM, LMU, and
coRNN---in several experiments that rely on long-range dependencies.
Each of the networks is based on a recurrent architecture, but with
different approaches to computing the weights and combining the
information across the temporal scales. Specifically, SITH builds a
log-compressed representation of the input signal and DeepSITH learns a
relevant set of features at every layer, which become the input to the
next layer. Where practical, we explicitly manipulate the characteristic
time scale of the experiment and evaluate the performance of the network
as the scale is increased.

We ran all of the experiments presented in
this work with the PyTorch machine learning framework \cite{Paszke.etal.2017}.
In Table \ref{tab:model_params} we show the hyperparameters
chosen for each experiment. We attempted to keep the number of learnable
parameters for each network to be close to those presented in the
current state of the art \cite{Stojnic.etal.2021}. For the following
experiments, we applied a 20\% dropout on the output of each DeepSITH
layer, except for the last one, during training. We utilized the Adam
optimization algorithm \cite{Kingma.Ba.2014} for training all of the
networks. Additionally, for all the following experiments except psMNIST, we
run each test 5 different times, and report the 95\% confidence intervals in
their respective results figures.

It should also be noted that we did not perform
a hyperparameter search for any networks a priori. For DeepSITH, we instead
followed the heuristics described above. For the three other networks, we used
hyperparameters presented in the paper's they were interoduced. In the cases
where hyperparameters were not explicitly published, we hand-titrated the values
of the hyperparameters based on ranges of values in each network's respective
publications. The LMU paper for the LMU and LSTM parameters, and the coRNN paper
for the coRNN parameters. If the publication did not have the same experiment,
we used the closest comparable experiment's presented hyperparameter ranges to
titrate for that network. All parameters for the compared network architectures
are provided in the suplemental materials.

\begin{table*}
 \caption{Parameter values of the DeepSITH network in each task.}
\begin{center}
\begin{sc}
\begin{small}
 \begin{tabular}{ c c c c c c c }
\toprule
 Experiment & \# Layers & $\tau_{max}$ & $k$ & \numtaustar{i} & \#hidden & Tot. Wts. \\
\midrule
 p/sMNIST & 3 & 30, 150, 750 & 125, 61, 35 & 20 & 60 & 146350 \\
\midrule
 Adding Problem & 4 & 20, 120, 720, 4320 & 75, 27, 14, 8 & 13 & 25 & 25151 \\
\midrule
 Mackey-Glass & 3 & 25, 50, 150 & 15, 8, 4 & 8 & 25 & 10301 \\
\midrule
 Hateful-8 & 4 & 25, 100, 400, 1200 & 35, 16, 9, 6 & 10 & 35 & 37808 \\
 \bottomrule
 \end{tabular}
 \label{tab:model_params}
\end{small}
\end{sc}
\end{center}
\vspace{-10pt}
\end{table*}

\hypertarget{permutedsequential-mnist}{%
\subsection{Permuted/Sequential MNIST}\label{permutedsequential-mnist}}

\begin{figure}
\hypertarget{fig:mnist_results}{%
\centering
\includegraphics[width=0.65\textwidth]{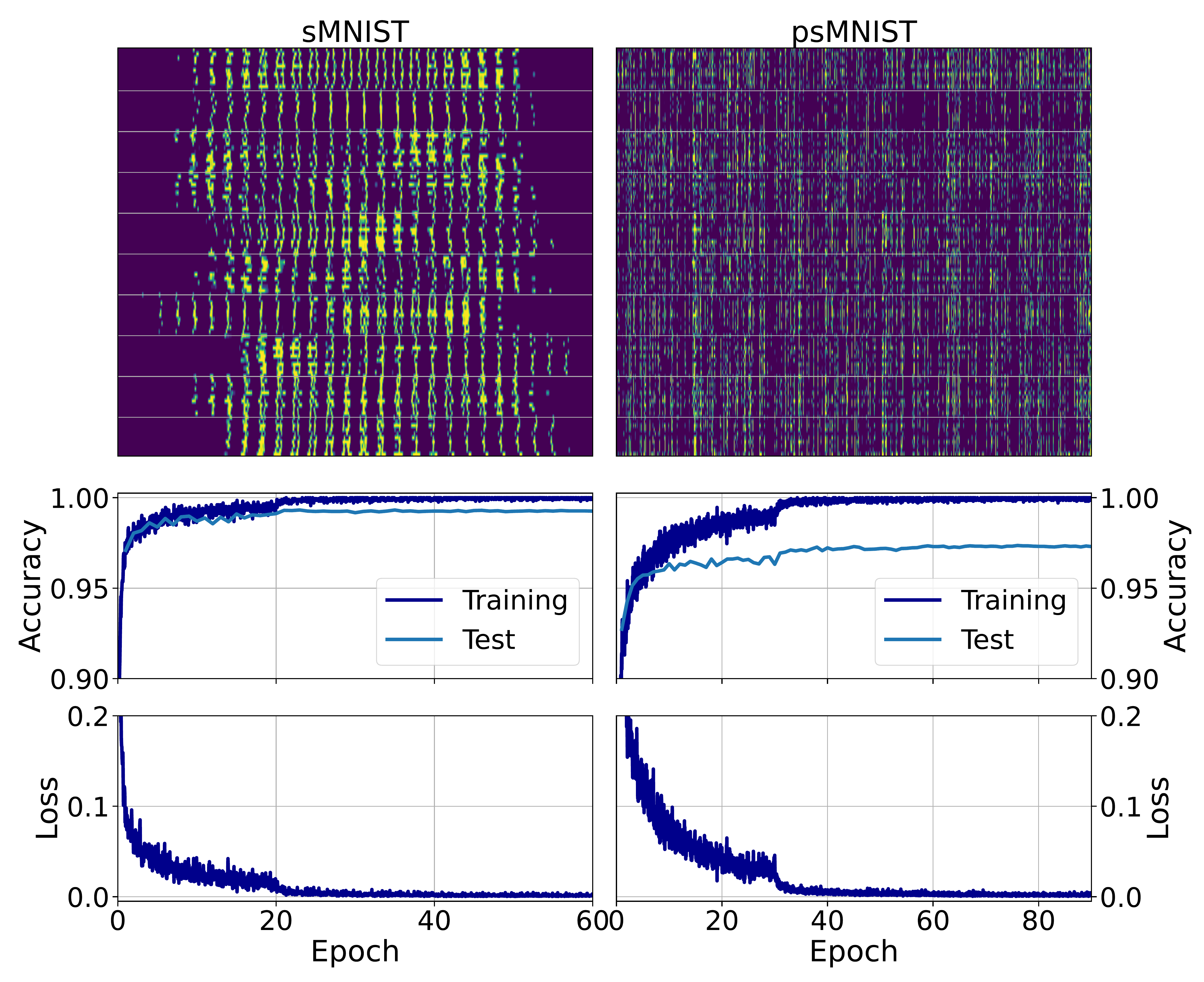}
\caption{\emph{The DeepSITH network achieves results comparable to state-of-the-art
on the sequential MNIST tasks.} \textbf{Top}:
Stacked in order are 10 examples of each of the 10 digits, flattened to
be 1-dimensional time series, from sMNIST and psMNIST items.
\textbf{Middle}: Plots the accuracy across epochs
on sMNIST and psMNIST. DeepSITH achieves a classification accuracy of
99.32\% on sMNIST, and 97.36\% on psMNIST with identically configured
networks. \textbf{Bottom}: Plots the loss across
epochs on sMNIST and psMNIST.}\label{fig:mnist_results}
}
\end{figure}

In the MNIST task \cite{Schwabbauer.1975}, handwritten numerical digits can
be identified by neural networks with almost 100\% accuracy utilizing a
convolutional neural network (CNN). This task is transformed into a more
difficult, memory intensive task by presenting each pixel in the image
one at a time, creating the time series classification task known as
sequential MNIST (sMNIST). An even harder task, permuted sequential
MNIST (psMNIST), is constructed by randomizing the order of pixels such
that each image is shuffled in the same way and then presented to the
networks for classification. Here, we train a DeepSITH network to
classify digits in the sequential and permuted sequential MNIST tasks by
learning to recognize multiple scales of patterns in each time series.

\begin{table}
 \caption{Results from DeepSITH on psMNIST and sMNIST
 compared to other networks. Best performances are in bold.}
\begin{center}
 \begin{tabular}{ c c c }
\toprule
 Network & psMNIST & sMNIST \\
\midrule
 DeepSITH & \textbf{97.36\%} & 99.32\% \\
\midrule
 LSTM & 90.20\% & 98.90\% \\
\midrule
 LMU & 97.15\% & N\/A \\
\midrule
 coRNN & 97.34\% & \textbf{99.40\%} \\
\bottomrule
 \end{tabular}
 \label{tab:mnist_res}
\end{center}
\vspace{-10pt}
\end{table}

The DeepSITH network is trained with a batch size of 64, with a
cross--entropy loss function, with a training/test split of 80\%-20\%.
In between each layer we applied batch
normalization, and applied a step learning rate annealing after every
third of the training epochs (2e-3, 2e-4, 2e-5). It should be noted that
the permutation that we applied in our tests was the same as in the
\cite{Rusch.Mishra.2020} study examining the coRNN, making the results
directly comparable. Test set accuracy was queried after every training
epoch for visualization purposes. While we did not attempt to minimize
the number of learnable weights in this network, DeepSITH had 146k
weights, compared to 134k, 102k, and 165k weights for the coRNN, LMU,
and LSTM networks, respectively.

The performance of DeepSITH on the standard (left) and permuted (right)
sequential MNIST task is shown in Figure \ref{fig:mnist_results}, along
with comparisons to other networks in Table \ref{tab:mnist_res}.
DeepSITH was able to achieve performance comparable to state-of-the-art on permuted
sequential MNIST with a test accuracy of 97.36\% \cite{Rusch.Mishra.2020},
and comparable performance to the best performing networks at 99.32\%
for normal sequential MNIST \cite{Li.etal.2019, Voelker.etal.2019}.
This result displays the expressivity of DeepSITH on this
difficult time series classification task.

\hypertarget{adding-problem}{%
\subsection{Adding Problem}\label{adding-problem}}

\begin{figure}
\hypertarget{fig:adding_results}{%
\centering
\includegraphics[width=0.9\textwidth]{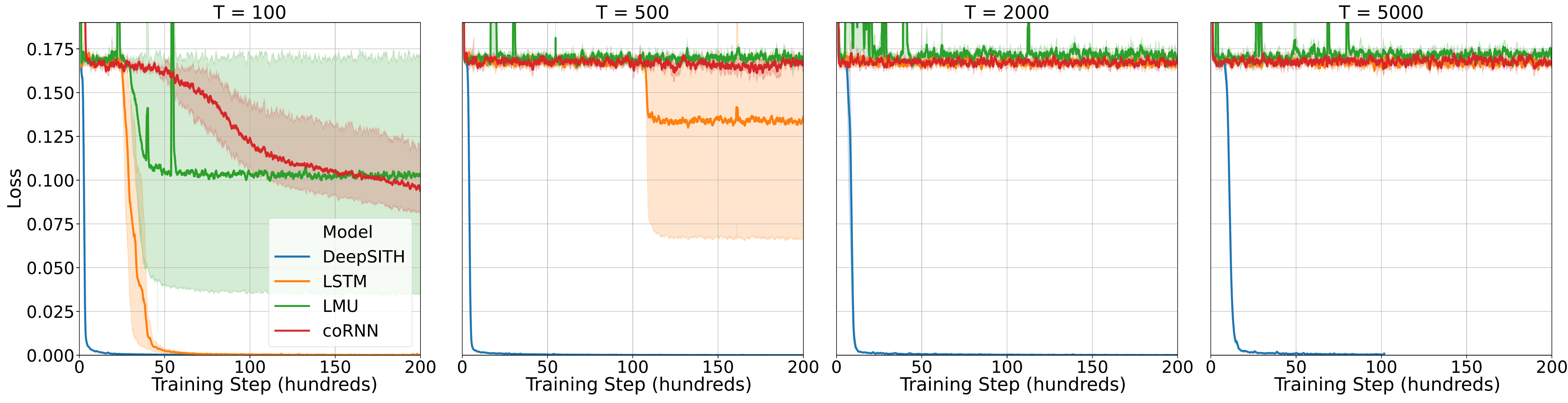}
\caption{\emph{DeepSITH learns the adding problem within the first 2500
training steps, faster than any other tested network.} Plotted here are
the running average mean squared error (MSE) losses over the previous 100
items trained with each network on signal lengths of \(T=100\),
\(T=500\), \(T=2000\), and \(T=5000\), over 5 runs with a 95\% confidence interval.
As \(T\) gets larger, the signals become much more difficult to learn, as they require the networks to
make associations between larger temporal
distances.}\label{fig:adding_results}
}
\end{figure}

We examined the Adding Problem \cite{Hochreiter.Schmidhuber.1997} as a
way to measure the ability of different networks to process and encode
long-range temporal associations. In this implementation of the task,
taken directly from \cite{Rusch.Mishra.2020}, a 2-dimensional time series
of length \(T\) is generated. The first dimension contains a series of
uniformly distributed values between 0 and 1, and the second dimension
contains all zeros except two indexes set to 1. These two indexes are
chosen at random such that one occurs within the first \(T/2\) indexes
and the other within the second \(T/2\) indexes. The goal of this task
is to maintain the random values from the first dimension presented at
the same time as the 1's in the second dimension until the end of the
sequence and then add them together. The evaluation criterion is mean
square error (MSE). The experiments were ran with a batch size of 50
items for all tests, and the loss was logged for every other batch. In
terms of complexity, DeepSITH had 25k learnable weights
for all sequence lengths, the LMU ranged from 2k to 11k , LSTM
had 67k across all lengths, and the coRNN had 33k across all lengths. For training
and testing, please note that all examples were randomly generated at every epoch.

Figure \ref{fig:adding_results} shows the results of the four networks
across different signal lengths, \(T\). The DeepSITH network learned the tasks
quickly at every length of signal without changing model parameters. As T
increases, the LMU and LSTM struggled with learning within a reasonable number
of training samples. The coRNN also struggled, but it should be noted that this
network has been shown in a previous study to learn the adding problem very well
at all signal durations tested here. We were unable to replicate the results
in any of the conditions. The DeepSITH network was able to solve the problem regardless
of signal duration.

\hypertarget{mackey-glass-prediction-of-chaotic-time-series}{%
\subsection{Mackey-Glass Prediction of Chaotic Time
Series}\label{mackey-glass-prediction-of-chaotic-time-series}}

The Mackey-Glass equations are a series of delay differential equations
originally applied to describe both healthy and abnormal variations in
blood cell counts and other biological systems \cite{Mackey.Glass.1977}.
Here we generated a one-dimensional time series with the differential
equations with different lag values, \(tau\), which controls a time scale of
the dynamics. The task was then to predict values some number of time
steps into the future as the time series is fed into the network.

\begin{wrapfigure}{l}{0.5\linewidth}
\hypertarget{fig:mg_results}{%
\centering
\includegraphics{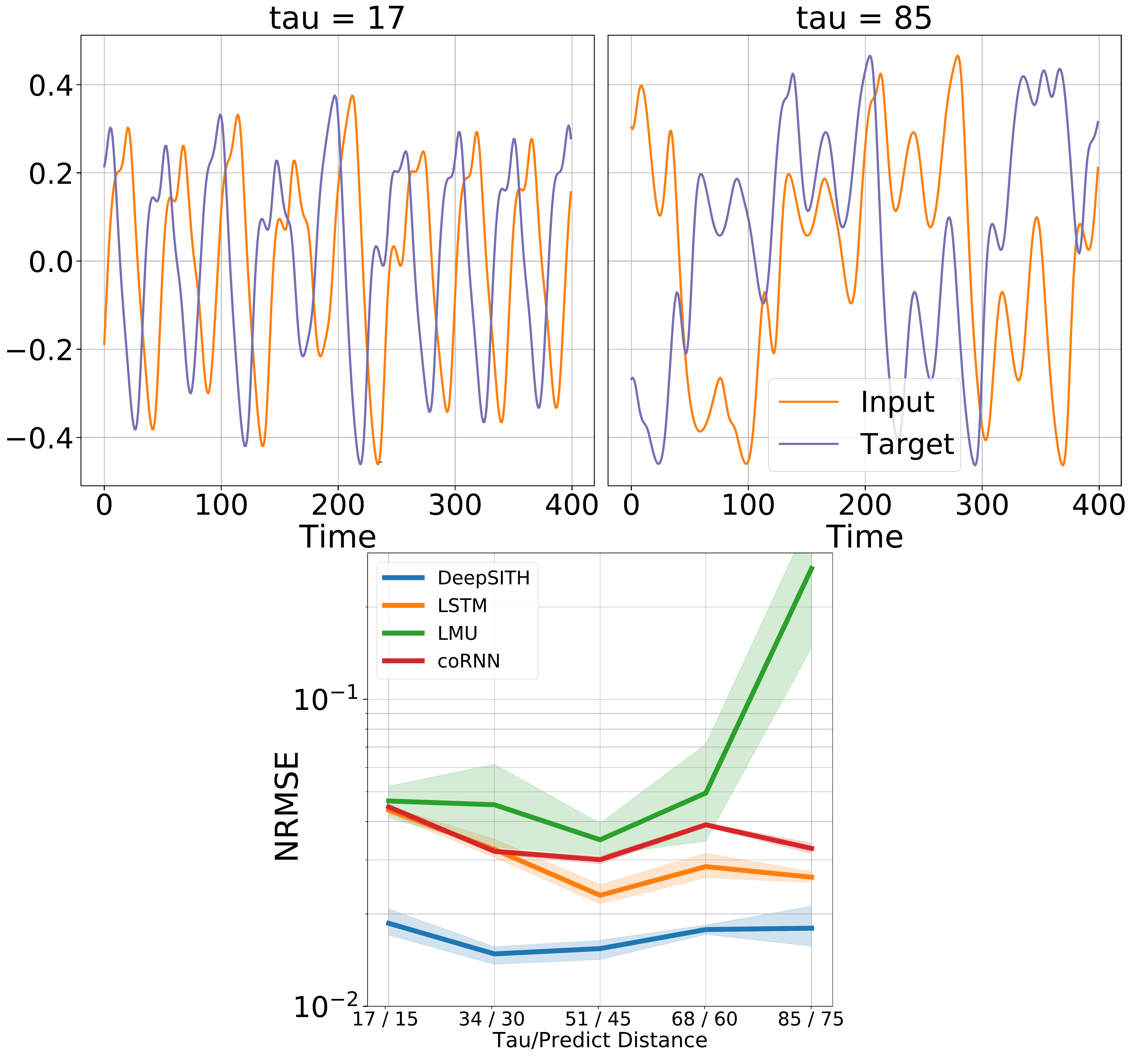}
\caption{\emph{DeepSITH learns to predict multiple levels of
Mackey-Glass complexity.} \textbf{Top} These plots contain examples of
Mackey-Glass time series with different values for \(tau\), which controls
the complexity of the signal. \textbf{Bottom} This plot contains
prediction results in terms of normalized root mean squared error for
different levels of signal complexity and number of timesteps into the
future the networks were predicting, 95\% confidence interval over 5 runs.}\label{fig:mg_results}
}
\end{wrapfigure}

The starting parameters for this experiment were taken from previous
work, with \(tau\) of 17 and the prediction distance into the future set to
15 time steps \cite{Voelker.etal.2019}. To test prediction accuracy for
increasing levels of complexity in the signal, we then generated signals
with increasing multiples of \(tau\), while also ensuring the ratio of
complexity to the prediction duration was kept constant, giving rise to
\(tau /\)prediction distance values of 17/15, 34/30, 51/45, 68/60, 85/75.
To illustrate the effect of \(tau\) on the complexity of the signal, we plot
examples for \(tau=17\) and 85 in the top of Figure \ref{fig:mg_results}.
Different values of \(tau\) introduce correlations at different temporal
scales into the chaotic dynamics, allowing us to test the networks'
ability to predict and encode multiple time-scales of information
simultaneously.

We generated 128 continuous one-dimensional signals for each value of
\(tau\), and split these signals for a 50\%-50\% training\\test split.
Each network was trained and tested separately on each
\(tau /\)prediction distance combination. Training was with a batch size
of 32, and the networks were evaluated on the testing set with normed
root mean squared error (NRMSE) calculated over predictions made at each
time step, as in previous work \cite{Voelker.etal.2019}. The
parameterizations for each network did not change across values of \(tau\).
For the LSTM, LMU, and DeepSITH, we kept the number of weights to around
18k, while the coRNN had 32k weights.

The bottom of figure \ref{fig:mg_results} shows the NRMSE for each
network as a function of \(tau\) and prediction distance over 5 runs. DeepSITH
outperforms the other architectures, and to our knowledge is performing at
state of the art on this task.

\hypertarget{the-hateful-8}{%
\subsection{The Hateful-8}\label{the-hateful-8}}

When learning to decode a time series after a delay, there may be noise
during the delay that will hinder the network's ability to maintain the
signal through time. Noise can be even more of an issue if it is similar
to the signal, such that it is difficult to separate signal from noise.
In such cases the network must learn the key features of the signal that
can enable successful classification, while ignoring similar features
until the end of the time series. Here, we introduce a novel time series
classification task based on Morse code, which we call the Hateful 8. In
Morse code, all letters are defined by a unique one-dimensional pattern
of dots (represented by activation lasting for one time step) and dashes
(represented by activation lasting for three time steps), each separated
by one inactivated time step, with three inactive time steps indicating
the end of a letter. In the Hateful 8 task there are 8 unique patterns
of dots and dashes making up the signals to decode, followed by noise
made up of semi-random dots and dashes similar to the signal.

\begin{figure}[t!]
\hypertarget{fig:h8_results}{%
\centering
\includegraphics[width=0.82\textwidth]{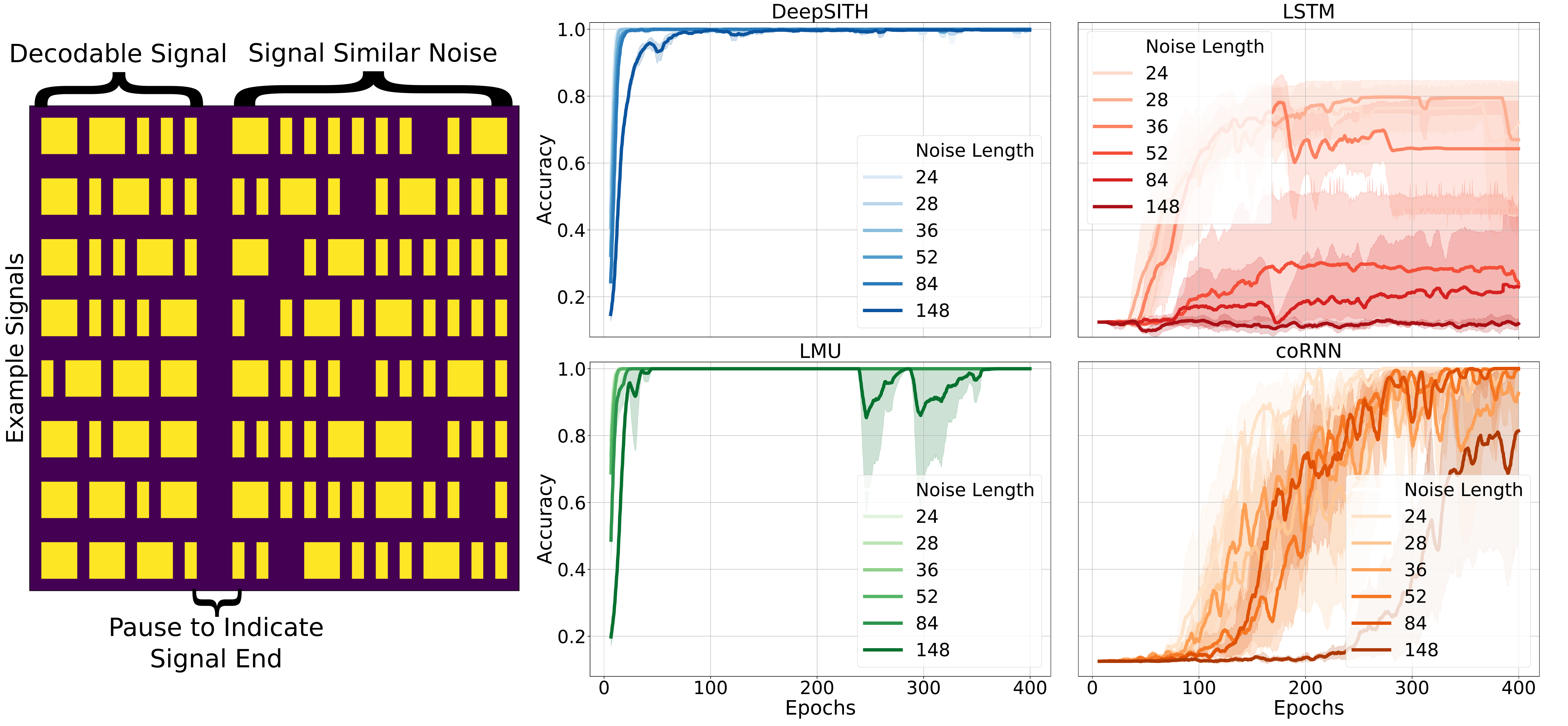}
\caption{\emph{DeepSITH consistently and quickly classifies long, noisy
time series signals known as the Hateful 8.} \textbf{Left} 8 example
Hateful-8 signals. The useful signal occurs in the first 17 time steps,
and after a short pause, the rest is semi-random, signal-similar noise.
\textbf{Center and Right} Plotted are the performance on a held out test
set after training for various noise lengths, over 5 runs with a 95\% confidence interval.
The darker color indicates longer noise durations. The DeepSITH and LMU
networks were able to learn this task the fastest, approximately 10 times
faster than the coRNN network.}\label{fig:h8_results}
}
\vspace{-15pt}
\end{figure}

The total decodable signal is 17 time steps long
(including the 3 time steps long pause), followed by signal-similar
noise. We trained and tested the four networks over increasingly longer
durations of noise added to the end of the decodable signal. Thus, to
exhibit high performance a network must maintain decoded information in
the face of noise that resembles the signal. We generated the
signal-similar noise pseudo randomly such that there is an even mix of
dots and dashes, along with pauses similar to the one to indicate the
end of the decodable signal. Figure \ref{fig:h8_results} plots examples
of all 8 different decodable signals followed by example signal-similar
noise.

For training, we generated 32 different versions of each of the Hateful
8, each with a different randomization of the noise. For the testing set, we again generated
randomized noise, and generated 10 noise patterns for each of the 8
different decodable portions. In total, we had 256 training signals and
80 test signals for each amount of noise. Note, each duration of noise was
trained and tested in independent experiments. Each network was trained and tested 5 times for each noise duration.
We trained networks with a batch size of 32. DeepSITH had 38k weights, LSTM had 30k weights, LMU had 30k weights, and
the coRNN 32k weights.

In the center and right of Figure \ref{fig:h8_results} we see the
performance of the four networks on the testing set as a function of
epoch and amount of noise. All networks exhibited some degree of success
with the lower amounts of noise, and only the LSTM was unable to learn
to 100\% accuracy at the higher noise lengths. In terms of training
time, the LMU and DeepSITH networks learned quickly and were more stable
than the coRNN and LSTM networks.

\hypertarget{discussion}{%
\section{Discussion}\label{discussion}}

DeepSITH is a brain-inspired approach to solving time series prediction
and decoding tasks in ML \cite{Fan.etal.2020}, especially those that
require sensitivity to temporal relationships among events. DeepSITH was
able to achieve near state-of-the-art performance in all the examples
examined in this paper. It is possible that some other set of hyper parameters
eould provide better results for DeepSITH or for the other networks tested.

For comparison within the p/sMNIST framework, we did
not have to select any hyperparameters for other networks since their
results were published in previous studies. With the p/sMNIST tasks,
DeepSITH was able to achieve near state-of-the-art
performance on permuted sMNIST and sMNIST proper. For Mackey-Glass prediction
DeepSITH achieved state of the art prediction accuracy. For the other networks, we used parameter
values directly from their source papers in the cases where they ran the
same experiments, otherwise we made an educated guess based on the
parameters in the published work. DeepSITH was able to learn the Adding
Problem with essentially the same amount of training time regardless of
the length of the input, and did so seemingly faster than the other
networks. It was also able to achieve 100\% accuracy on the Hateful-8
problem regardless of how much noise we presented, as was the LMU
network. We attribute the successes of DeepSITH to the scale-invariance of the
memory representation and the decomposition of what information from
when information.

We compared DeepSITH to LMU \cite{Voelker.etal.2019} and coRNN \cite{Rusch.Mishra.2020},
which are both recent approaches based on modified RNNs.
They both demonstrated remarkable success in time series prediction on
similar problems, paving the road for new approaches that tackle the
problem differently from most common RNNs, such as LSTM. While similar
in spirit, LMU and coRNN use different memory representations than the
one proposed here.

The compression introduced by \(\Phi_k(t/\taustar)\) results in a loss
of information about timing of events as they recede into the past. On
its face, this may seem like a disadvantage of this approach. However,
the ubiquity of logarithmic scales in perception and psychophysics
suggest that this form of compression is adopted widely by the brain
\cite{Howard.2018,GuoEtal21, CaoEtal21}.
Viewed from one perspective, this gradual loss of information is a
positive good. The blur induced by \(\Phi_k\) forces the system to
generalize over a range of time points. The width of that range scales
up linearly with the time point in the past at which the observation is
made. The deep architecture with SITH representations enables the
network to store fine temporal relationships at early layers and turn
them into features that are retained over the entire range of time scales at
the next layer.  The temporal representation at each layer is
logarithmically-compressed but extending over a progressively larger  range of
scales.

The ubiquity of logarithmic scales in perception and neuroscience raises the
question of what adaptive benefit has led to this form of compression.
Logarithmic scaling may be a reaction to power law statistics in the natural
world \cite{WeiStoc12}.   Perhaps the choice of logarithmic scale allows the
organism to provide an equivalent amount of information about environments
with a wide variety of intrinsic scales \cite{HowaShan18}.  Another
possibility, suggested by recent work \cite{JansLind21, JacqEtal21a} is that
the choice of a logarithmic scale enables a perceptual network to generalize
to unseen scales.  For instance, consider a convolutional filter
trained on a time series projecting onto a representation of $\log $
time.  Rescaling time $t \rightarrow at$, that is presenting the time series
faster or slower, has the effect of translating the representation over $\log$
time by $\log a$.  Because convolution is translation-covariant, a CNN would
(ignoring edge effects) identify the same features, but at translated
locations in $\log$ time.  Thus logarithmic compression could allow deep
networks to show to perceptual invariances to changes in scale.   This
property could enable CNNs to learn faster and generalize across a wider range
of stimuli.

\hypertarget{societal-impact}{%
\section{Societal Impact}\label{societal-impact}}

Given that we are introducing a methodology that is attempting to function as a
drop-in replacement for RNNs, we also inherit the fundamental risks associated
with improving the accuracy of these types of network architectures.
Action-detection, voice recognition, and other time-series
decoding tasks run the risk of being heavily biased, and by introducing a human
inspired method we run the risk of introducing more human-like biases.
Furthermore, in the realm of natural language processing we run the risk of
creating an AI that better captures long-range dependencies and is,
consequently, able to generate extremely human-like text, which could
potentially add to the already problematic fake-news epidemic. Extra caution
should be taken that these methods are steered away from applications that
could be used maliciously, but we believe that, ultimately, these AI advances
will do more good than harm.

\section{Acknowledgements}

This material is based upon work supported by the Defense Advanced Research
Projects Agency (DARPA) under Agreement No. HR00112190036. The authors
acknowledge Research Computing at The University of Virginia for providing
computational resources and technical support that have contributed to the
results reported within this publication. URL: https://rc.virginia.edu

\bibliography{SITHCon}

\section*{Checklist}

\begin{enumerate}

\item For all authors...
\begin{enumerate}
  \item Do the main claims made in the abstract and introduction accurately reflect the paper's contributions and scope?
    \answerYes{See Section \ref{discussion}}
  \item Did you describe the limitations of your work?
    \answerYes{Briefly, See Section \ref{discussion}}
  \item Did you discuss any potential negative societal impacts of your work?
    \answerYes{See Section \ref{societal-impact}}
  \item Have you read the ethics review guidelines and ensured that your paper conforms to them?
    \answerYes{}
\end{enumerate}

\item If you are including theoretical results...
\begin{enumerate}
  \item Did you state the full set of assumptions of all theoretical results?
    \answerNA{}
	\item Did you include complete proofs of all theoretical results?
    \answerNA{}
\end{enumerate}

\item If you ran experiments...
\begin{enumerate}
  \item Did you include the code, data, and instructions needed to reproduce the main experimental results (either in the supplemental material or as a URL)?
    \answerYes{See Supplementals: Github}
  \item Did you specify all the training details (e.g., data splits, hyperparameters, how they were chosen)?
    \answerYes{See Table \ref{tab:model_params} \& Section \ref{experiments}}
	\item Did you report error bars (e.g., with respect to the random seed after running experiments multiple times)?
    \answerNo{We report only the best performing models}
	\item Did you include the total amount of compute and the type of resources used (e.g., type of GPUs, internal cluster, or cloud provider)?
    \answerNo{We did not ran anything that could not be ran on a single GPU}
\end{enumerate}

\item If you are using existing assets (e.g., code, data, models) or curating/releasing new assets...
\begin{enumerate}
  \item If your work uses existing assets, did you cite the creators?
    \answerYes{We used MackeyGlass item generation code, several models, and the MNIST dataset. See \ref{experiments}}
  \item Did you mention the license of the assets?
    \answerNo{}
  \item Did you include any new assets either in the supplemental material or as a URL?
    \answerNA{}
  \item Did you discuss whether and how consent was obtained from people whose data you're using/curating?
    \answerNA{}
  \item Did you discuss whether the data you are using/curating contains personally identifiable information or offensive content?
    \answerNo{It does not have personally identifiable information or offensive content.}
\end{enumerate}

\item If you used crowdsourcing or conducted research with human subjects...
\begin{enumerate}
  \item Did you include the full text of instructions given to participants and screenshots, if applicable?
    \answerNA{}
  \item Did you describe any potential participant risks, with links to Institutional Review Board (IRB) approvals, if applicable?
    \answerNA{}
  \item Did you include the estimated hourly wage paid to participants and the total amount spent on participant compensation?
    \answerNA{}
\end{enumerate}

\end{enumerate}

\end{document}